\documentclass[10pt,twocolumn,letterpaper]{article}

\usepackage{btas}
\usepackage{times}
\usepackage{epsfig}
\usepackage{graphicx}
\usepackage{amsmath}
\usepackage{amssymb}

\graphicspath{{./images/}}
\usepackage{url}

\btasfinalcopy % *** Uncomment this line for the final submission

\ifbtasfinal\fi

\begin{document}

%%%%%%%%% TITLE
\title{Iris and Periocular Recognition in Arabian Race Horses Using\\ Deep Convolutional Neural Networks}

\author{Mateusz Trokielewicz$^{\dag,\ddag}$\\
$^{\dag}$Biometrics Laboratory\\
Research and Academic Computer Network\\
Kolska 12, 01-045 Warsaw, Poland\\
$^{\ddag}$Institute of Control and Computation Engineering\\
Warsaw University of Technology\\
Nowowiejska 15/19, 00-665 Warsaw, Poland\\
{\tt\small mateusz.trokielewicz@nask.pl}
\and
Mateusz Szadkowski$^{\star}$\\
$^{\star}$Department and Clinic of Animal Surgery\\
Faculty of Veterinary Medicine\\
University of Life Sciences\\
Gleboka 30, 20-612 Lublin, Poland\\
{\tt\small matszadkowski@gmail.com}
}

\maketitle
\thispagestyle{empty}

\begin{abstract}
This paper presents a study devoted to recognizing horses by means of their iris and periocular features using deep convolutional neural networks (DCNNs). Identification of race horses is crucial for animal identity confirmation prior to racing. As this is usually done shortly before a race, fast and reliable methods that are friendly and inflict no harm upon animals are important. Iris recognition has been shown to work with horse irides, provided that algorithms deployed for such task are fine-tuned for horse irides and input data is of very high quality. In our work, we examine a possibility of utilizing deep convolutional neural networks for a fusion of both iris and periocular region features. With such methodology, ocular biometrics in horses could perform well without employing complicated algorithms that require a lot of fine-tuning and prior knowledge of the input image, while at the same time being rotation, translation, and to some extent also image quality invariant. We were able to achieve promising results, with EER=9.5\% using two network architectures with score-level fusion.\let\thefootnote\relax\footnote{Accepted for publication at the IEEE IJCB 2017, Denver, USA}
\end{abstract}

\section{Introduction}
\label{sec:Intro}
\subsection{Horse identification}
Identification plays a vital role in handling, breeding, and trading horses. Still in use persists the word-based and graphical description of a horse, which is obviously prone to errors and subjectivity. Until very recently, popular methods of identifying horses included applying burnt or frozen marks on the skin -- which can be considered harmful, invasive, and animal hostile. Tattoos placed on the inside of the upper lip are also widely in use. As of today, the most popular methods incorporate electronic tagging, with wireless RFID transponders being implanted deeply into the neck muscles on the left side. The device is identified with a numerical code consisting of 15 digits. The horse is also given a unique \emph{Universal Equine Life Number}, also comprising 15 digits containing vital information. These data are also placed in the animal's passport required for trading. 

In cases of pure breed horses, a lot of attention is given to confirming the genetic origin of a horse. This is possible thanks to genetic tests, which are also reported in the passport. Despite the availability of this document for each horse, fast identification usually involves reading the RFID transponder. This leaves possibility of falsification, as the RFID tag can be replaced - an achievement not easy and probably leaving a scar, yet certainly not impossible.

\subsection{Iris recognition and periocular recognition}
Iris recognition, a biometric method of identification persons, utilizes rich individual features found in the texture of the iris. Uniqueness and low genotype dependence of these traits, combined with effective methodologies of extracting and comparing features enable accurate and fast identification of persons. A typical iris recognition system, such as the one proposed \cite{Daugman1993} and originally patented \cite{DaugmanPatent} by John Daugman, employs iris image segmentation with pupil and iris being approximated with circles, followed by a feature extraction based on bandpass filtering such as Gabor wavelets, output phase quantization and binary code creation, and finally comparison of the two codes using exclusive-OR (XOR) operation to produce a dissimilarity metric in a form of fractional Hamming distance from the range [0, 1], with values close to zero expected for same-eye (\emph{genuine}) comparisons, and values close to 0.5 expected for different-eye (\emph{impostor}) comparisons. 

Periocular biometrics, on the other hand, employs features of the entire eye region, such as the location and shape of eye corners, eyelids, eyelashes, eyebrows, fragments of the nose, \etc. Compared with iris recognition, periocular recognition often requires less constrained image acquisition conditions, such as imaging in visible light, at-a-distance, or on-the-move. However, these methods are far less capable of delivering excellent recognition accuracy. 

Our contribution is thus to make use of both iris and periocular features for horse identification. Employing convolutional neural networks for this task seems like a natural approach, as they are capable of taking a whole, unprocessed image as an input and and predict a class label by hierarchical feature extraction and classification. To our knowledge, this is the first study to employ DCNNs for horse recognition. Also, a novel database is offered to interested researchers for non-commercial purposes. This paper is structured as follows. Section \ref{sec:Related} discusses past work on animal identification using ocular biometrics and convolutional neural networks. Section \ref{sec:Anatomy} provides a description of the anatomy of equine eye. Database of equine eye images collected by the authors is detailed in Section \ref{sec:Database}. Section \ref{sec:Networks} contains implementation details for the networks used for classification, with experiments and results reported on in Section \ref{sec:Experiments}. Finally, Section \ref{sec:Conclusions} gives the conclusions, current limitations of this study, and future research plans.

\section{Related work}
\label{sec:Related}

\subsection{Recognizing animals by their ocular traits}
When it comes to iris recognition in horses, or in large animals in general, not many peer-reviewed publications are available. Suzaki \etal \cite{SuzakiHorseIris2001} proposed a complete system for recognizing horses via their iris patterns. Their solution included a complicated segmentation procedure that approximates the iris with non-oval shapes and takes only the lower portion of the iris into consideration, to account for the noise that may be introduced by \emph{granula iridica}, a melanin pigment concentration that is usually visible near the upper pupil-iris boundary of a horse iris. Gabor convolution with filtering output quantization is then employed for the encoding stage. 

More work has been done regarding farm animals identification.  Zhang \etal \cite{ZhangIrisLocalizationCows2009} propose a cow iris localization method comprising two steps. First, a threshold transform, Sobel operator for edge detection, and inner boundary fitting with quadratic B-spline interpolation are employed to provide initial, coarse localization results. This establishes priors for fine tuning, which is carried out with non-concentric circles fitting. Following that, an outline for a full iris recognition system for cattle identification, employing the same method for image segmentation and 2D Gabor filtering and amplitude quantization for iris encoding was proposed by Wang \etal \cite{WangIrisLargeAnimals2009}. Sun \etal \cite{SunSIFTIrisCowsElsevier2013} aimed at implementing reliable iris recognition in cattle, reporting on several issues and challenges, such as large variance of image rotation and scaling, non-circular irides and pupils, off-angle presentations and multiple light reflections. Their approach uses active contours for image segmentation and scale-invariant feature transform (SIFT) for iris feature extraction. Then, histogram representations are generated using bag-of-features, and finally the matching stage is carried out by calculating distances between these histograms. Correct recognition rate of 98.15\% is reported. Recently, another solution for cow identification system employing iris biometrics was introduced by Lu \etal \cite{LuIrisCows2014}, utilizing edge detection and ellipse fitting for segmentation and two-dimensional Complex Wavelet Transform for iris encoding. Authors claim their segmentation procedure to be faster and more accurate than state-of-the-art methods developed for human iris segmentation, such as active contours and Hough transform, while at the same time being notably faster. 
Interestingly, Crouse \etal \cite{LemurRecognition2017} created a system for recognizing lemurs to aid preserving endangered species. A modified human face recognition system, based on patch-wise multi-scale LBP features and adjusted facial image normalization techniques, is said to achieve a correct recognition rate of 98.7\%.
 
 Notably, that all of the above methods require large \emph{a priori} knowledge regarding the object of recognition and fine-tuning of many parameters to obtain a well-performing solution. In this study, we aim at exploiting the flexibility that DCNNs offer, together with ease of implementation in an end-to-end manner with unprocessed images taken as input and class classification as model output.   
 
\subsection{Deep Convolutional Neural Networks}
Convolutional Neural Networks (CNNs), and especially Deep CNNs, utilize a two-dimensional convolution operation to learn image features at progressively higher-order scales, in turn being able to characterize both fine, texture-related and coarse, more abstract and high-level attributes of the input. These solutions have recently been shown to achieve excellent performance in tasks such as image classification, such as ZF-Net \cite{ZFNet2013}, AlexNet \cite{AlexNet2014}, VGG-Net \cite{VGGSimonyanCNNsForRecognition2014}, ResNet \cite{ResNet2013}, and image segmentation: pixelwise image labeling \cite{LiPixelwiseClassificationCNNs2014} and \cite{PinheiroPixelWiseLabelingCNNs2015}, semantic segmentation  \cite{YuCNNsForSemanticSegmentationARXIV2016}\cite{WuCNNsForSemanticSegmentation2014}, dense image labeling  \cite{IslamDenseImageLabelingCNNs2016}, employing deconvolutions to retrieve dense output \cite{NohDeconvolutionNNsForSegmentation2015}. 

Regarding DCNNs applications in ocular biometrics, Liu \etal \cite{LiuICB2016CNNsForIrisSegmentation} employed multi-scale fully convolutional networks for the purpose of segmenting noisy iris images (\eg visible light images with light reflections, blurry images captured on-the-move and/or at-a-distance, 'gaze-away' eyes, \emph{etc.}), with iris pixels being located without any \emph{a priori} knowledge or hand-crafted rules. The authors reported good performance of their solution, with error rates below 1\% of disagreeing pixels between the CNN approach and manual annotation. Minaee \etal \cite{MinaeeCNNforIrisRec2016} studied the possibility of extracting iris features with a convolutional network and passing them into a multi-class SVM classifier. The solution is said to achieve up to 98\% correct recognition rate on the IIIT Delhi database. Zhao \etal \cite{ZhaoKumarCNNsForPeriocularTIFS2016} studied periocular recognition by means of improved convolutional neural networks enhanced over classical CNN approaches by using additional models trained specifically to discriminate against features other that those person-specific, \eg skin tone, gender, left/right eye (soft biometrics, here referred to as \emph{semantic information}). In addition to the main network trained by identity, they also train an additional model to differentiate between males and females and between left and right eyes. Features from both networks then are reduced in dimensionality using PCA and fused together, with similarity score predicted using the joint Bayesian metric. The experiments were performed on several publicly available databases, including UBIRIS.v2 (EER=10.98\% obtained in a verification scenario using only the identity-discriminant model) and CASIA.v4-distance (EER=6.61\%, same). The semantics-assisted CNN, comprising both models was used in an identification setting, reaching 82.43\% and 98.90\% Rank-1 hit rate on the two above databases, respectively.   

\section{Horse eye anatomy}
\label{sec:Anatomy}
\subsection{Eye region of a horse}
\begin{figure}[!htb]
\centering
\includegraphics[width=\columnwidth]{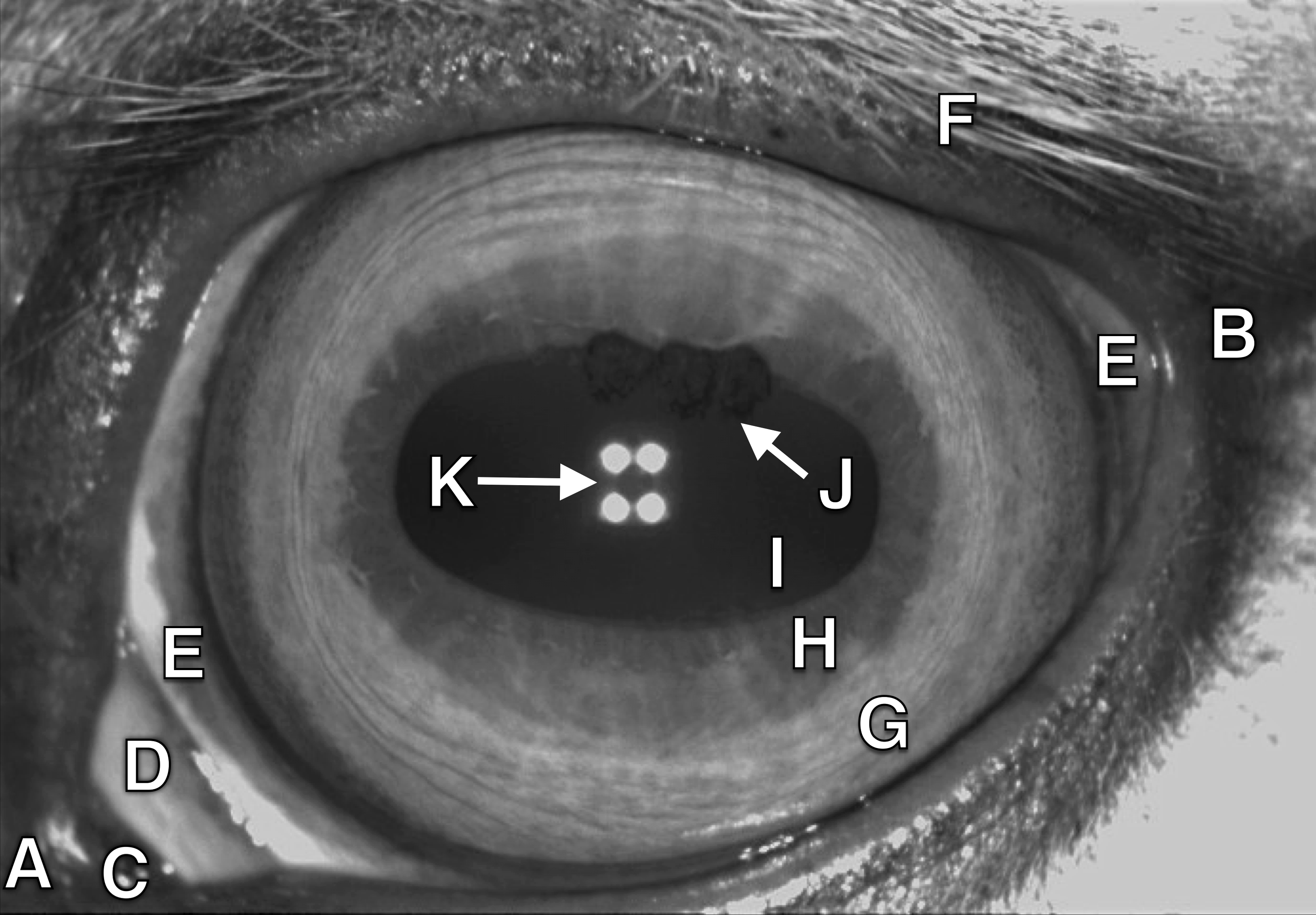}
\vskip0.2cm
\caption{Surface anatomy of equine eye and adnexa photographed under near-infrared illumination: (A) medial canthus, (B) lateral canthus, (C) lacrimal caruncle, (D) free margin of membrana nictitans, (E) sclera, (F) cilia, (G) ciliary zone of iris, (H) pupillary zone of iris, (I) pupil, (J) granulae iridica, (K) infrared light source reflection.}
\label{fig:horse-eye}
\end{figure}

Fig. \ref{fig:horse-eye} depicts the anterior segment of the horse's eyeball under near-infrared illumination. The palpebral fissure is horizontally oval. The eyelashes on the upper eyelid are well developed, but absent on the lower eyelid. Thick sensing hairs (\emph{vibrissae}) are located on the base of the eyelids. Near the medial canthus there is the lacrimal caruncle of varying size. Between the globe and the medial canthus there is a free margin of the third eyelid (\emph{membrana nicitans}). There is a relatively small portion of exposed sclera which is often pigmented at the lateral part. The largest visible portion of the eyeball surface is the cornea, which is elliptical in shape with regular, smooth, shiny surface. The horizontal axis of cornea is longer than the vertical axis. Transparency of the cornea is attributed to anatomical factors, such as the absence of blood vessels, nonkeratinized surface, lack of pigmentation and specific organization of stromal collagen fibers. Behind the cornea is the anterior chamber, filled with translucent aqueous humor \cite{GilgerEquineOphthalmology2005}\cite{GelatVeterinary2007}.
\subsection{Equine iris}
The iris separates the anterior chamber of the globe from the posterior chamber. It extends centrally from the ciliary body and partially covers the anterior surface of the lens. A central aperture of the iris constitutes the pupil. Dilation and constriction of the pupil is determined by the amount of light entering the eyeball. In horses, the pupil is rod-shaped with a long horizontal axis. Along the dorsal (upper) pupillary margin there are well developed round black masses called \emph{granula iridica} (\emph{corpora nigra}). Sometimes, similar features can exist on the ventral (lower) edge of the pupil. Anterior iris surface is composed of two zones, namely the central pupillary zone and the peripheral ciliary zone. The pupillary zone is often both darker and thinner than the rest of the iris. Color of the iris varies from dark brown to gold, or even blue, and it depends on the amount of iridal stroma pigmentation, the type of pigmentation, vascularization, and on the shape and size of collagen fibers found in the stroma.  The original surface shape of iris is determined by a specific system of blood vessels and collagen fibers \cite{GelatVeterinary2007}.

\begin{figure}[!h]
\centering
\includegraphics[width=0.75\columnwidth]{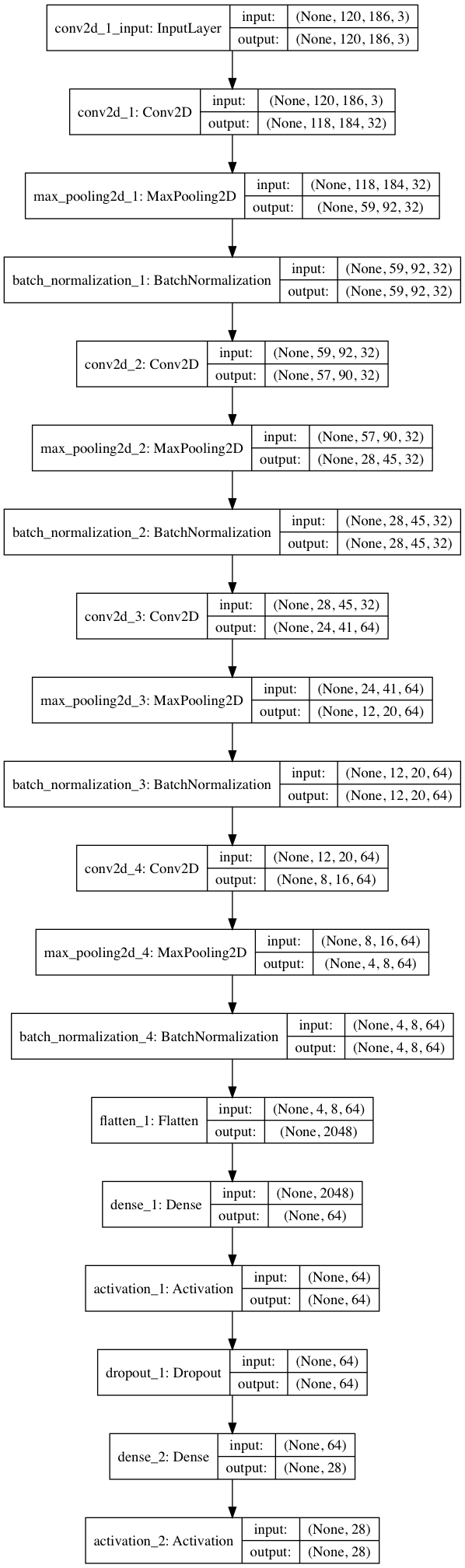}
\vskip0.2cm
\caption{Architecture of the HorseNet-4 model.}
\label{fig:architecture-4}
\end{figure}

\section{Database of horse eye images}
\label{sec:Database}
\textbf{Collection protocol and equipment.} During the course of this study, we have collected a novel database of images representing eyes of horses. Twenty eight animals had their eyes photographed: 14 mares (50\%), 10 stallions (35.7\%) and 4 geldings (14.3\%) aged 1 to 24 (with an average of 9 years old). The largest group were Arabian horses (15 animals, 53.6\%). A typical ophthalmological examination with TonoVet, slit lamp, PanOptic, BPI 50 and Retinomax 3 equipment was performed in all horses prior to iris image acquisition. In one animal, due to pathological changes in the cornea and in the anterior chamber, biometric data were collected unilaterally from the remaining clinically healthy eyeball. For iris image collection horses had their eyes photographed using a specialized veterinary device Pupil-LR (by SIEM Bio-Medicale) originally purposed for ophthalmological examinations, but at the same time capable of producing high-quality near-infrared photographs of the eye and its surroundings. Sample image obtained from the device is shown in Fig. \ref{fig:horse-eye}. The device outputs 480x744 grayscale images. The dataset we created contains eye images belonging to 28 distinct classes. Note that \textbf{class} here relates to a single animal, not to a single animal eye. This is due to the fact that images of left and right \textbf{eye} of the same horse are expected to exhibit partial dependence on each other, in contrast to \textbf{iris} texture, which is believed to be independent between eyes of the same individual. Two acquisition sessions separated by a few minutes were performed, each class being represented by approx. 2000 images.

\textbf{Censoring.} As it is impossible to make the horse stand still without movement and blinking during such a period of time, the resulting material was inspected manually for photos representing only the eyelids or those severely out of focus. Such samples were excluded from the final dataset.

\textbf{Data variants.} Following initial censoring, the data was screened for the second time to create two versions of it: one containing all samples available, and one additionally inspected to remove samples of imperfect quality, such as images slightly out of focus or those missing small portions of the eye. We later refer to those sets as \textbf{high quality} and \textbf{mixed quality} databases, respectively. The latter dataset is represented by approximately 20\% more samples than the former. This is done to assess whether to train the networks using data of mixed quality, or should it be pre-screened to achieve better results.

\textbf{Database access.} For the purposes of reproducibility and to encourage further research, the database collected by the authors as a part of this study is available to all interested researchers for non-commercial use. See: \emph{http://zbum.ia.pw. edu.pl/EN$\rightarrow$Research$\rightarrow$Databases}

\section{Convolutional networks implementation}
\label{sec:Networks}
Two variants of a deep convolutional neural network has been constructed for the task: HorseNet-4 and HorseNet-6, with four and six convolutional layers, respectively.

\begin{figure}[!h]
\centering
\includegraphics[width=0.67\columnwidth]{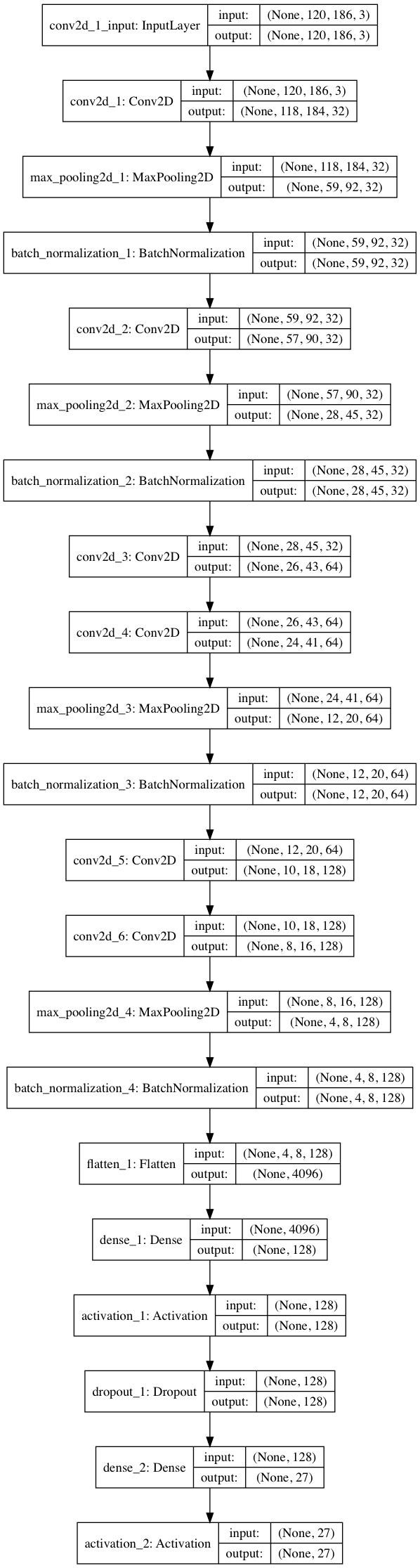}
\vskip4mm
\caption{Architecture of the HorseNet-6 model.}
\label{fig:architecture-6}
\end{figure}

\textbf{Architecture.} The network receives an unmodified (apart from $4\times$ downsampling) input eye image, which is then processed through several convolutional layers (\emph{Conv2D}) with a $3\times3$ and $5\times5$ kernels, each of them followed by a max pooling layers (\emph{MaxPooling2D}) with a pool size of $2\times2$ with stride of 1. Next, a fully-connected layer (\emph{Dense1}) is put after the last \emph{Conv2D+MaxPooling2D+BatchNormalization} set of layers. Finally, a second \emph{Dense2} layer with \emph{Softmax} activation function outputs the estimated probability that the input sample belongs to one of $N$ classes, where $N$ denotes the number of animals. Architectures of the HorseNet-4 and HorseNet-6 networks are shown in detail in Figs. \ref{fig:architecture-4} and \ref{fig:architecture-6}.

\textbf{Batch normalization.} Batch normalization is applied after each pooling layer. This method enables accelerated learning of the model by normalizing training mini-batches, and in turn alleviating the changes in the distribution of each layer's inputs (reducing \emph{internal covariate shift}) \cite{BatchNormalization}, providing1 faster convergence of the network.
   
\textbf{Dropout.} Dropout was applied after the first \emph{Dense1} layer, with a probability of dropping any given unit set to $p=0.5$. Dropout is a technique used to fight against network overfitting due to neuron co-adaptation (units reliance on other units) and employs random removal of a selected portion of neurons and their connections during training to prevent over-adaptation of the network \cite{Dropout}.

\textbf{Training.} The model was trained using the Stochastic Gradient Descent (SGD) optimizer with learning rate of 0.01, decay of $1e^{-6}$ and momentum of 0.9. The training was performed using all samples from session 1, being parsed to the network in \emph{mini-batches} of 32. 9 epochs were selected as a number that prevents network over-fitting.

\textbf{Evaluation.} Evaluation was performed by feeding testing samples (all samples from session 2) through the network and taking the highest output from the \emph{Softmax} layer.

\textbf{Technical details.} The implementation of the network was performed with a Python \cite{Python} + OpenCV \cite{opencv_library} environment, using the Keras framework for deep learning \cite{Keras} with TensorFlow backend \cite{tensorflow2015}. All calculations were done on an Intel Core i5 notebook running at 2.7 GHz. Training of the larger model takes approximately 25-35 minutes per epoch, depending on the training methodology, while the average prediction time per sample is 0.03 seconds.

\section{Experiments and results}
\label{sec:Experiments}
\subsection{Performance on different datasets}
\begin{figure}[!t]
\centering
\includegraphics[width=\columnwidth]{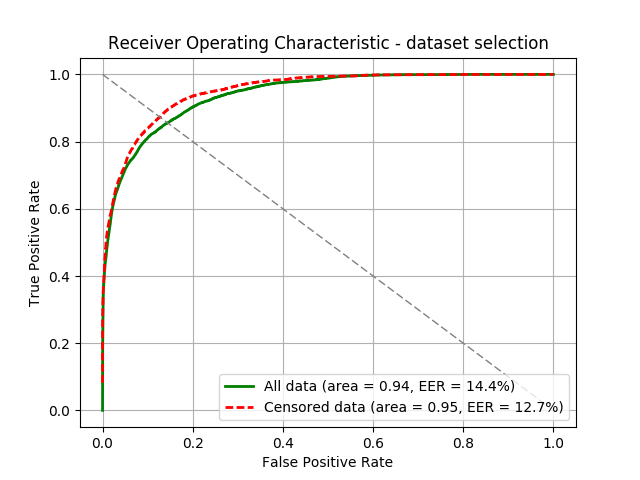}
\vskip0.2cm
\caption{Performance on datasets of different quality. Unilateral recognition methodology and HorseNet-4 architecture is used.}
\label{fig:roc-datasets}
\end{figure}

In the first experiment we evaluate whether it is better to train the network on the original dataset, which contains images slightly out of focus and images not containing an entire region of the eye, or to train it on the additionally censored dataset, which includes only images of highest quality. Fig. \ref{fig:roc-datasets} presents the ROC curves obtained for both datasets. The testing was performed on session 2 samples, which were not used for the training. Censored dataset lets us achieve better accuracy (EER of 12.7\% versus 14.4\%).

\subsection{Unilateral vs bilateral recognition}
\begin{figure}[!t]
\centering
\includegraphics[width=\columnwidth]{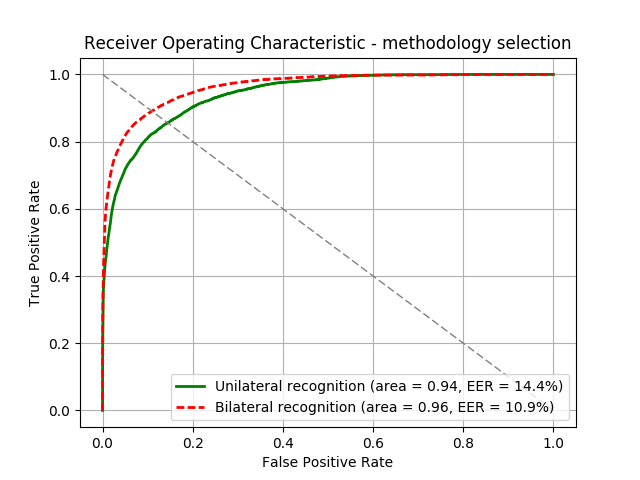}
\vskip0.2cm
\caption{Performance of different approaches to network training: unilateral vs bilateral. Uncensored data is used.}
\label{fig:roc-methodology}
\end{figure}

The second experiment deals with answering the following question: is it better to use only one eye (unilateral recognition), or both eyes of the animal (bilateral recognition)? The former implies enrollment of a left or right eye of a given horse, while the in the latter approach we enroll both eyes of an animal and give them the same label. According to ROC graphs (Fig. \ref{fig:roc-methodology}), bilateral variant perform better, with EER=10.9\% achieved on the HorseNet-4. 

\subsection{Network architecture selection}
\begin{figure}[!t]
\centering
\includegraphics[width=\columnwidth]{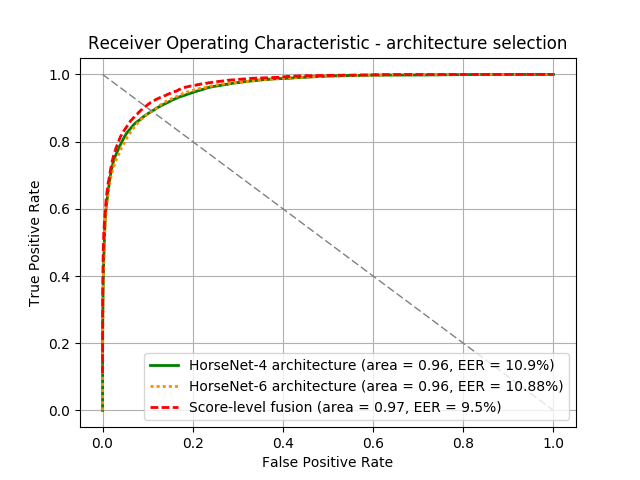}
\vskip0.2cm
\caption{Performance of different models: HorseNet-4 vs HorseNet-6 vs score-level fusion approach. Censored data is used.}
\label{fig:roc-architectures}
\end{figure}

Finally, with the censored dataset and bilateral recognition methodology, we examine the performance of the network architectures and of their combination with score-level fusion. Here, both models perform comparatively (EER=10.9\% and EER=10.88\% for HorseNet-4 and HorseNet-6, respectively), however, fusing the two scores by averaging the \emph{Softmax} outputs returns a significantly better accuracy with EER=9.5\%, Fig. \ref{fig:roc-architectures}. Interestingly, an uncensored dataset performs better, probably due to the fact that the larger number of training examples is able to compensate for slightly worse data quality and enables better learning of the eye features. 

\section{Conclusions}
\label{sec:Conclusions}
This paper gives valuable insight into deploying deep convolutional neural networks for the purpose of horse recognition. Experiments regarding several questions about the approach details, such as dataset censoring, identification methodology and network architectures are presented. We managed to show, with encouraging, yet not perfect results, that horse recognition with DCNNs using ocular biometrics may be possible. Notably, the database collected by the authors is made publicly available.

We are aware that this study is restrained by some limitations. Difficulties may arise when attempting fast acquisition of horse eye images due to movement of the animal. A specialized device is used here, so future experiments should focus on employing more popular cameras for this purpose -- mobile phones, for instance. Also, the procedure of recognition described above is a closed-set identification, so the networks have to be re-trained every time a new horse appears. Nonetheless, the results presented above are certainly promising and point to a possibility of employing this emerging group of methods for the purpose of identifying race horses. Applications in human iris and periocular recognition is also imaginable.

\section*{Acknowledgements}
The authors would like to thank Dr. Adam Czajka and Prof. Andrzej Pacut for their invaluable remarks and edits that contributed to the quality of this paper. We are also indebted to Prof. Ireneusz Balicki for providing the equipment necessary to proceed with the examination of horse irides.   

{\small
\bibliographystyle{ieee}
\bibliography{refs}
}

\end{document}